%% file: main.tex
\documentclass{article}
\pdfoutput=1

\usepackage{spconf}

\usepackage{graphicx}
\graphicspath{{fig/}}

\usepackage{enumitem}
\usepackage{microtype}
\usepackage{flushend}
\usepackage{url}
\usepackage{tabularx}
\usepackage{multirow}
\newcommand{\mytable}{\centering
                      \renewcommand{\arraystretch}{1.2}
                      }
\newcolumntype{C}{>{\centering\arraybackslash}X}
\newcolumntype{L}{>{\raggedright\arraybackslash}X}

\usepackage{amsmath}
\usepackage{amssymb}
\renewcommand{\vec}[1]{\boldsymbol{\mathbf{#1}}}

\usepackage{cite}  

\title{Deep convolutional acoustic word embeddings \\ using word-pair side information \vspace*{-5pt}
} 
\name{Herman Kamper$^1$, Weiran Wang$^2$, Karen Livescu$^2$\thanks{This research was supported by NSF grant IIS-1321015.  The opinions expressed in this work are those of the authors and do not necessarily reflect the views of the funding agency. HK is funded by a Commonwealth Scholarship.} \vspace*{-9pt}}
\address{$^1$CSTR and ILCC, School of Informatics, University of Edinburgh, UK \\
         $^2$Toyota Technological Institute at Chicago, USA \\
         {\small \tt h.kamper@sms.ed.ac.uk, weiranwang@ttic.edu, klivescu@ttic.edu} \vspace*{-6pt}}

\begin{document}
\ninept

\maketitle

\begin{abstract}
Recent studies have been revisiting whole words as the basic modelling unit in speech recognition and query applications, instead of phonetic units. Such whole-word segmental systems rely on a function that maps a variable-length speech segment to a vector in a fixed-dimensional space; the resulting \textit{acoustic word embeddings} need to allow for accurate discrimination between different word types, directly in the embedding space. We compare several old and new approaches in a word discrimination task. Our best approach uses side information in the form of known word pairs to train a Siamese convolutional neural network (CNN): a pair of tied networks that take two speech segments as input and produce their embeddings, trained with a hinge loss that separates same-word pairs and different-word pairs by some margin. A word classifier CNN performs similarly, but requires much stronger supervision. Both types of CNNs yield large improvements over the best previously published results on the word discrimination task.
\end{abstract}
\vspace*{-0.35\baselineskip}
\vspace*{-0.05\baselineskip}

\input{intro}
\input{embedding}
\input{exp}
\input{conc}

\newpage
\bibliographystyle{IEEEbib}
\bibliography{mybib}

\end{document}

%% file: intro.tex
\section{Introduction}
\label{sec:introduction}

Most current speech processing systems rely on a deep architecture to classify speech frames into subword units (often phone states).
This approach still relies on frame-level independence assumptions as well as a pronunciation lexicon for breaking up words into their subword constituents.
As an alternative, some 
researchers~\cite{dewachter+etal_taslp07,heigold+etal_icassp12,levin+etal_asru13,bengio+heigold_interspeech14,guoguo+etal_icassp15,maas+etal_icmlwrl12,rasanen_cogsci15}
have started to reconsider using whole words as the basic modelling unit.

Some of the earliest speech recognition systems were based on template-based whole-word modelling~\cite{myers+rabiner_tassp81}.  This idea has been revisited in modern template-based automatic speech recognition (ASR) systems~\cite{dewachter+etal_taslp07,heigold+etal_icassp12}, as well as modern speech indexing applications such as query-by-example search~\cite{zhang+glass_asru09,zhang+etal_icassp12}.
These systems typically use dynamic time warping (DTW) to quantify the similarity of phone or word segments of variable length.
Recent work has also considered frame-level embeddings which map acoustic features to a new frame-level representation that is tailored to word discrimination when combined with DTW~\cite{jansen+etal_icassp13b,kamper+etal_icassp15,thiolliere+etal_interspeech15}.
DTW, however, has known inadequacies~\cite{rabiner+etal_tassp78} and is 
quadratic-time in the duration of the segments.

Levin {\it et al.}~\cite{levin+etal_asru13} 
proposed a segmental approach where an arbitrary-length speech segment is embedded in a fixed-dimensional space such that segments of the same word type have similar embeddings. Segments can then be compared by simply calculating a distance in the embedding space, a linear time operation in the embedding dimensionality.
Several approaches were developed in~\cite{levin+etal_asru13}, and in~\cite{levin+etal_icassp15} these were successfully applied in a query-by-example search system.

Bengio and Heigold~\cite{bengio+heigold_interspeech14} similarly 
used whole-word fixed-dimensional representations in a segmental 
ASR lattice rescoring system.
Their acoustic embeddings are obtained from a convolutional neural network (CNN), trained 
with a combination of a word classification and a ranking loss.
When combining the hypotheses of the baseline system with the embedding-based scores, ASR performance was improved.
A similar approach was followed in~\cite{guoguo+etal_icassp15},
where long short-term memory (LSTM) 
networks were used to obtain whole-word embeddings for a 
query-by-example search task.
Finally, Maas {\it et al.}~\cite{maas+etal_icmlwrl12} trained a regression CNN that reconstructs a semantic word embedding from acoustic speech input; these features were used in a segmental conditional random field ASR system.

In this paper we compare 
several CNN-based approaches to each other and to the best approach of Levin {\it et al.}~\cite{levin+etal_asru13}, on a word discrimination task.
This task has been used in several other studies~\cite{carlin+etal_icassp11,jansen+etal_icassp13b,kamper+etal_icassp15} to assess the accuracy of acoustic embedding approaches without the need to train a complete recognition 
or search system.
Building on
ideas from earlier CNN-based approaches, we propose new networks 
that make use of weaker supervision in the form of known word pairs.
The approach is based on \textit{Siamese networks}: tied networks that take in pairs of 
input vectors and minimize or maximize a distance depending on whether a pair comes from the same or different 
classes~\cite{bromley+etal_ijpr93}.
We show that a Siamese CNN trained with with a hinge-like contrastive loss function outperforms the best approach of Levin {\it et al.}~\cite{levin+etal_asru13}, and performs similarly to a word classifier CNN, despite the weaker form of supervision.
By reducing the Siamese CNN embedding dimensionality with a post-processing linear discriminant analysis, we also obtain a more compact embedding that maintains best performance.

%% file: embedding.tex
\vspace{-.05in}
\section{Acoustic word embedding approaches}
\vspace{-.05in}

\begin{figure*}[!t]
    \centering
    {\footnotesize(a)} \includegraphics[scale=0.775]{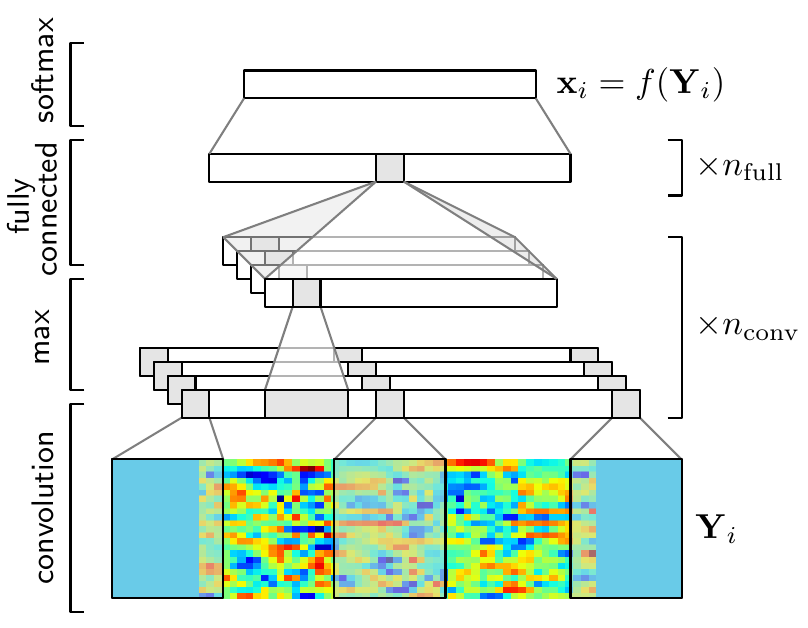} \hspace{3pt}
    {\footnotesize(b)} \includegraphics[scale=0.775]{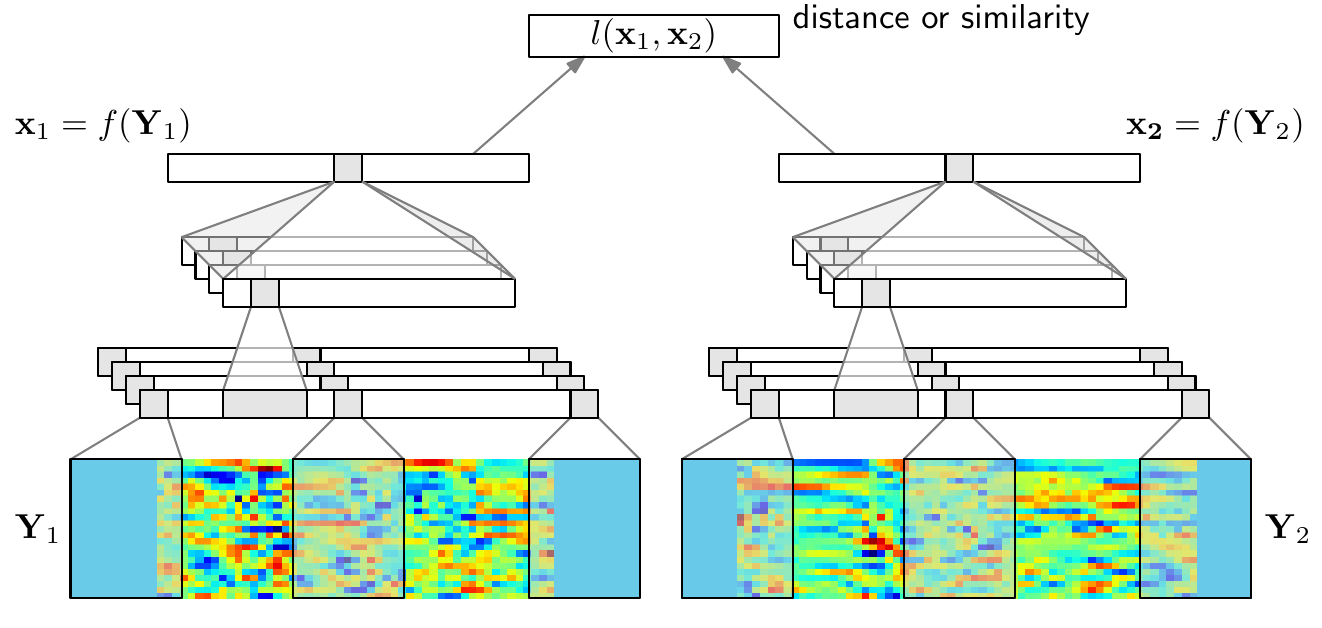} \\
    \vspace*{-0.7\baselineskip}
    \caption{\textup{(a)} Word classification CNN and \textup{(b)} word similarity Siamese CNN for obtaining acoustic word embeddings from padded speech input.}
    \label{fig:cnns}
    \vspace*{-0.9\baselineskip}
\end{figure*}

For speech applications using fixed-dimensional representations of whole words, it is desirable to find a mapping such that word segments of the same type are close in the embedding space while those of different types are far from each other.
Formally, we use the notation $Y = \vec{y}_{1:T}$ to denote a vector time series, where each $\vec{y}_t \in \mathbb{R}^b$ is 
a $b$-dimensional frame-level
feature vector (e.g.~MFCCs).
An acoustic word embedding approach
is a function $f(Y)$ that maps arbitrary-length time series $Y$ into a fixed-dimensional space $\mathbb{R}^d$;
if $Y_1$ and $Y_2$ are two word segments, the distance between the vectors $f(Y_1)$ and $f(Y_2)$ should indicate whether they are of the same word~type~or~not.

Typical embedding approaches use a training set of known word segments $\mathcal{Y}_{\text{train}} = \{ Y_i \}_{i = 1}^{N_{\textrm{train}}}$ to learn $f$.
Different degrees of supervision can be assumed, ranging from unsupervised, where the only knowledge of $\mathcal{Y}_{\text{train}}$ is that it contains unidentified word segments, to supervised, where the word label for each segment is known.
Below we review previous work (Sections~\ref{sec:reference_vector} and~\ref{sec:word_classification}) and then present our 
own approaches which use weak supervision in the form of known word pairs (Section~\ref{sec:word_embeddings}), and can also additionally use word labels to find lower-dimensional but still accurate embeddings (Section~\ref{sec:dimred}).

\subsection{Reference vector methods}
\label{sec:reference_vector}

Several embedding approaches were proposed and compared in Levin {\it et al.}~\cite{levin+etal_asru13} based on 
the idea of using a 
{\it reference vector}
to construct the mapping $f$.
For a target speech segment, a reference vector 
consists of the DTW alignment cost to every exemplar in a reference set $\mathcal{Y}_{\text{ref}} \subseteq
\mathcal{Y}_{\text{train}}$.
Applying dimensionality reduction to the reference vector yields the desired embedding in $\mathbb{R}^d$.
The intuition is that the content of a speech segment should be characterized well through its similarity to other segments.  Such embeddings have subsequently been used for keyword search~\cite{levin+etal_icassp15} and unsupervised term discovery~\cite{kamper2015fully}.
One drawback is the need to compute a large number of DTW alignments.
Several dimensionality reduction approaches were considered in~\cite{levin+etal_asru13}, and in Section~\ref{sec:results} we compare to their best overall approach which uses a combination of Laplacian eigenmaps (a non-linear graph embedding approach) and linear discriminant~analysis.

\subsection{Word classification CNN}
\label{sec:word_classification}

For the whole-word speech recognition system in~\cite{bengio+heigold_interspeech14}, Bengio and Heigold proposed that, when word labels $\mathcal{W}_{\text{train}} = \{ w_i \}_{i = 1}^{N_{\textrm{train}}}$ are available for the training segments $\mathcal{Y}_{\text{train}}$, a supervised neural network can simply be trained to predict the word class (type) given the speech as input.
The softmax prediction layer of such a neural network then gives a fixed-dimensional representation in $\mathbb{R}^d$, where $d$ here is the number of distinct word types (the vocabulary size).
During testing, some inputs 
may correspond to unseen words, but even in these cases the softmax layer gives a fixed-dimensional distributional representation of the input in terms of seen word~types.

A standard feed-forward neural network classifier, however, requires fixed-dimensional input.
A simple solution was used in~\cite{bengio+heigold_interspeech14}: 
All word segments are padded to the same length, given by the maximum duration of a word segment in $\mathcal{Y}_{\text{train}}$.
Instead of using fully-connected layers,
convolutional 
and pooling layers are used to alleviate the effect of the padding.
A convolutional neural network (CNN) such as this is shown in Figure~\ref{fig:cnns}(a).
Our implementation uses mean-normalized MFCCs which are zero-padded to $n_\text{pad}$ frames.
One-dimensional convolution is performed only over time, covering a number of frames and all features, and is followed by max-pooling.
These layers can be repeated.
A number of fully-connected layers is used next, which feeds into the final softmax layer.
Formally, the whole CNN defines a mapping function $f : \vec{Y}_i \in \mathbb{R}^{b \times n_\text{pad}} \rightarrow \vec{x}_i \in \mathbb{R}^d$, that takes input $\vec{Y}_i$, obtained by padding the variable-length $b$-dimensional vector time series~$Y_i$, and produces the acoustic embedding $\vec{x}_i$.

Instead of using the representation from the word classification CNN directly, Bengio and Heigold~\cite{bengio+heigold_interspeech14} used a paired network with a ranking loss to map acoustic word embeddings into a common space with orthography-based word embeddings obtained by also mapping the word labels $\mathcal{W}_{\text{train}}$ into a lower-dimensional space.
This was done to make it possible to use the classifier outputs in a particular lattice rescoring architecture, which requires scores for lattice arcs.
The evaluation framework we use (Section~\ref{sec:setup}) is designed to be decoupled from a recognition architecture, and we can therefore use the distributional representation from the classifier CNN directly. 
An investigation of whether embeddings of word labels can be additionally used to improve acoustic word embeddings is left for future~work.

\vspace*{-0.5\baselineskip}
\subsection{Word similarity Siamese CNNs}
\label{sec:word_embeddings}

If the labels $\mathcal{W}_{\textrm{train}}$ for the training set $\mathcal{Y}_{\textrm{train}}$ are not known, a weaker form of supervision that has also been used~\cite{jansen+etal_icassp13b,kamper+etal_icassp15,renshaw+etal_interspeech15,thiolliere+etal_interspeech15} is the knowledge that pairs of word segments in $\mathcal{Y}_{\textrm{train}}$ share the same unknown word type: $\mathcal{S}_{\textrm{train}} = \{ (m, n): (Y_m, Y_n) \textrm{ are of the same type} \}$.
This type of side information is appealing since it is often easier to obtain in low-resource settings, for example by using an unsupervised term discovery system~\cite{park+glass_taslp08,jansen+vandurme_asru11} to find unidentified matching word pairs.

Such paired supervision has been used for several 
problems and domains, including phonetic discovery~\cite{badino+etal_icassp14,thiolliere+etal_interspeech15}, semantic word 
embeddings~\cite{huang+etal_cimk13,mikolov+etal_arxiv13,wieting+etal_tacl15} and vision applications~\cite{hadsell+etal_cvpr06}.
Many of these studies use \textit{Siamese networks}, a term used since the early 
1990s to describe a pair of networks with tied parameters which is trained to 
optimize a distance function between representations of two data instances~\cite{bromley+etal_ijpr93}.
To train these networks it is sometimes assumed that pairs not in $\mathcal{S}_{\textrm{train}}$ belong to different types; we also make this assumption here.

Figure~\ref{fig:cnns}(b) illustrates how we apply this idea to obtain acoustic word embeddings.
The two sides of our Siamese network take padded inputs $\vec{Y}_1$ and $\vec{Y}_2$.
For the two sides we use CNNs similar to that of the word classification CNN.
But instead of terminating in a softmax layer, the final fully connected layer on each side gives the desired acoustic embedding.
In initial experiments on development data, we considered several loss functions, and here we focus only on the most successful ones.
We found that losses based on cosine similarity outperformed Euclidean-based losses, and in particular the coscos$^2$ loss from~\cite{synnaeve+etal_slt14} gave the best performance of the losses in~\cite{bromley+etal_ijpr93,synnaeve+etal_slt14}:
\vspace*{-0.2\baselineskip}
\begin{equation}
    l_{\text{cos cos}^2}(\vec{x}_1, \vec{x}_2) =
    \begin{cases}
        \frac{1 - \cos(\vec{x}_1, \vec{x}_2)}{2} & \text{if same} \\
        \cos^2(\vec{x}_1, \vec{x}_2) & \text{if different} \\
    \end{cases}
    \label{eq:coscos2}
\vspace*{-0.2\baselineskip}
\end{equation}
This loss pushes the angle between embeddings of the same type to be zero, while embeddings of different types are pushed to be~orthogonal.

In discrimination tasks, the decision of whether two data instances are of the same type is not based on their \textit{absolute} distance, but rather their \textit{relative} distance compared to other pairs.
This 
motivates a 
margin-based (hinge) loss, similar to that of~\cite{mikolov+etal_arxiv13,wieting+etal_tacl15}: 
\vspace*{-0.3\baselineskip}
\begin{equation}
    l_\text{cos hinge} = \max \left\{ 0, m + d_\text{cos}(\vec{x}_1, \vec{x}_2) - d_\text{cos}(\vec{x}_1, \vec{x}_3) \right\}
    \label{eq:coshinge}
\end{equation}
where $ d_\text{cos}(\vec{x}_1, \vec{x}_2) = \frac{1 - \cos(\vec{x}_1, \vec{x}_2)}{2}$ is the cosine distance between $\vec{x}_1$ and $\vec{x}_2$, and $m$ is a margin parameter.
Here, $\vec{x}_1$ and $\vec{x}_2$ are always of the same type while $\vec{x}_1$ and $\vec{x}_3$ are of different types.
This loss is therefore at a minimum when all embeddings $\vec{x}_1$ and $\vec{x}_2$ of the same type are more similar by a margin $m$ than embeddings $\vec{x}_1$ and $\vec{x}_3$ of different types.
The margin also gives some leeway to the model.

Although Siamese networks have been used widely, to our knowledge this is the first work which uses Siamese networks (in particular Siamese CNNs) to obtain acoustic word embeddings from~speech.

\subsection{Controlling embedding dimensionality}
\label{sec:dimred}

We aim to learn word embeddings that are both discriminative and compact (low-dimensional).  The desired dimensionality may be guided by both computational and data constraints, and we may wish to be able to adjust it.
For word classification networks (Section~\ref{sec:word_classification}), the output dimensionality is given by the vocabulary size.  In our experiments (next section), we explore adjusting the dimensionality by inserting an additional linear bottleneck layer before the final softmax, with the number of units corresponding to the desired final dimensionality.  In Siamese networks (Section~\ref{sec:word_embeddings}) the final dimensionality can be directly tuned. 
If we have access to word labels $\mathcal{W}_{\textrm{train}}$ in addition to word pairs $\mathcal{S}_{\textrm{train}}$, we can also perform additional dimensionlity reduction on the Siamese CNN outputs using a supervised technique; in our experiments we use linear discriminant analysis (LDA).

%% file: exp.tex
\section{Experiments}
\label{sec:experiments}

\subsection{Evaluation and experimental setup}
\label{sec:setup}

Ultimately we would like to evaluate the different acoustic embedding approaches for downstream speech recognition and search tasks.
However, we do not want to be tied to a specific recognition 
architecture, and we would like to quickly compare many embedding approaches.
We therefore use a 
word discrimination task developed for this purpose~\cite{carlin+etal_icassp11}; in the \textit{same-different task}, we are given a pair of acoustic segments, each corresponding to a word, and we must decide whether the segments are examples of the same or different~words.

This task can be approached in a number of ways, but typically it is done either via a DTW score between segments (when using frame-by-frame embeddings), or via a Euclidean or cosine distance between vectors (when embedding complete segments).  In our evaluation, after training a model on $\mathcal{Y}_{\textrm{train}}$, the acoustic word embeddings of a disjoint test set $\mathcal{Y}_{\textrm{test}}$ are 
computed.
For every word pair in this set, the cosine distance\footnote{We also tried Euclidean distance, but 
as in~\cite{levin+etal_asru13,kamper+etal_icassp15,thiolliere+etal_interspeech15}, cosine worked~better.}
is calculated between their embeddings.
Two words can then be classified as being of the same or different type
based on some threshold, and a precision-recall curve is obtained by
varying the threshold.
To evaluate embeddings across different operating points, the area under the precision-recall curve is calculated
to yield the final evaluation metric, referred to as the average precision (AP).

We use data from the Switchboard corpus of English conversational telephone speech.
Data is parameterized as Mel-frequency cepstral coefficients (MFCCs) with first and second order 
derivatives, yielding 39-dimensional feature vectors.
Cepstral mean and variance normalization (CMVN) is applied per conversation side.
For the training set $\mathcal{Y}_{\textrm{train}}$ we use the set of about 10k word tokens from~\cite{jansen+etal_icassp13b,kamper+etal_icassp15};
it consists of word segments of at least 5 characters and 0.5 seconds in duration extracted from a forced alignment of the transcriptions, and comprises about 105 minutes of speech.
For the Siamsese CNNs, this set results in about 100k word segment pairs for $\mathcal{S}_{\textrm{train}}$.
For testing, we use 
the 11k-token set $\mathcal{Y}_{\textrm{test}}$ from~\cite{levin+etal_asru13,jansen+etal_icassp13b,kamper+etal_icassp15}, making the results from these studies directly comparable to the results obtained here.\footnote{In~\cite{levin+etal_asru13}, a slightly different training set was used. 
Nevertheless, the size of their training set is 
comparable to the set used here.}
This set was extracted from a portion of Switchboard distinct from $\mathcal{Y}_{\textrm{train}}$.
Similarly, we extracted an 11k-token development set. 

As mentioned in Section~\ref{sec:introduction}, recent studies~\cite{jansen+etal_icassp13b,kamper+etal_icassp15} have also been using frame-level embedding approaches in combination with DTW to perform the same-different task.
These approaches map the original features to a new frame-level representation that is tailored to word discrimination.
We compare our results to that of~\cite{jansen+etal_icassp13b}, which uses posteriograms over a partitioned universal background model (UBM), as well as~\cite{kamper+etal_icassp15}, which uses a correspondence autoencoder.

\subsection{Network architectures}

We used the Theano~\cite{bergstra+etal_scipy10} toolkit to implement the CNN-based models of Sections~\ref{sec:word_classification} and~\ref{sec:word_embeddings}.\footnote{CNN code:  {\scriptsize \url{https://github.com/kamperh/couscous}}. Complete recipe: {\scriptsize \url{https://github.com/kamperh/recipe_swbd_wordembeds}}.}
Models are trained using 
ADADELTA~\cite{zeiler_arxiv12}, an adaptive learning rate 
stochastic optimization method that adapts over time based on an accumulation of past gradients; we set the momentum hyper-parameter to $\rho = 0.9$ and the precision parameter to $\epsilon = 10^{-6}$.
Input speech segments are padded to $n_{\text{pad}} = 200 \text{ frames}$ (2~s), which corresponds to the longest word segment in $\mathcal{Y}_{\text{train}}$.
The architectures of the CNNs were optimized separately on the development data for each network type, resulting in the following structures:
\begin{itemize}[leftmargin=\parindent,itemsep=1pt,parsep=0pt,partopsep=0pt,topsep=1pt]
    \item \textit{Word classifier CNN:} 1-D convolution with 96 filters over 9 frames; ReLU; max pooling over 3 units; 1-D convolution with 96 filters over 8 units; ReLU; max pooling over 3 units; 1024-unit fully-connected ReLU; softmax layer over 1061 word types.
    \item \textit{Word similarity Siamese CNN:} two convolutional and max pooling layers as above; 2048-unit fully-connected ReLU; 1024-unit fully-connected linear linear; terminates in loss $l(\vec{x}_1, \vec{x}_2)$.
\end{itemize}
For the word classifier CNN, we only train on words in $\mathcal{Y}_{\text{train}}$ that occur at least three times; this gives a subset of 87\% of all tokens with 1061 unique word types.  {This minimum count was tuned on the development set.}  
To see the effect of the convolutions, we also train a word classifier deep neural network (DNN) using two 2048-unit fully-connected ReLU layers and a 1061-unit softmax layer.
For the Siamese CNN using $l_\text{cos hinge}$, we use a margin $m = 0.15$ 
(tuned on the development set).
If we had used ReLUs in the final layer in the Siamese CNNs, the angles between embeddings would be restricted to $[0, \pi/2]$; we therefore use a final linear layer.
All weights are initialized randomly; we run all models with five different initializations and report average performance and standard~deviations.

\subsection{Results}
\label{sec:results}

\begin{table}[!t]
    \vspace*{-0.5\baselineskip}
    \mytable
    \caption{Average precision (AP) on the test set.
    Models 1 to 3: best prior DTW-based approaches; model 4: best prior acoustic word embedding approach based on reference vectors; models 5 to 11: this work.
    For models 1 to 3, dimensionality (dim.) is at the frame level; for models 4 to 11 it is the segment embedding dimensionality.}
    \vspace*{4pt}
    \begin{tabularx}{\linewidth}{|@{\,}r@{\,}|L|r||l|}
        \hline
        \textbf{\#} & \textbf{Representation} & \textbf{Dim.} & \textbf{AP} \\
        \hline \hline
        1 & MFCCs with CMVN          & 39 & 0.214 \\
        2 & Best partitioned UBM~\cite{jansen+etal_icassp13b}                  & 100           & 0.286 \\
        3 & Correspondence autoencoder~\cite{kamper+etal_icassp15}        & 100           & 0.469 \\
        \hline
        4 & Reference vector approach~\cite{levin+etal_asru13}                       & 50            & 0.365 \\
        \hline
        5 & Word classifier DNN                                                     & 1061          & $0.301 \pm 0.005$ \\
        6 & Word classifier CNN                                                     & 1061          & $0.532 \pm 0.014$\\
        7 & ------ & 50 & $0.474 \pm 0.012$ \\
        8 & Siamese CNN, $l_{\text{cos cos}^2}$ loss                          & 1024         & $0.342 \pm 0.026 $ \\
        9 & Siamese CNN, $l_{\text{cos hinge}}$ loss                           & 1024         & $\textbf{0.549} \pm 0.011$\\
        10 & ------ & 50 & $0.504 \pm 0.011$ \\
        11 & LDA on model~9                                                  & 100          & $0.545 \pm 0.011 $ \\
        \hline
    \end{tabularx}
    \label{tbl:results}
    \vspace*{-0.9\baselineskip}
\end{table}

Table~\ref{tbl:results} shows AP performance on the test set from previous studies (models 1 to 4) as well as our newly implemented models (5 to 11).

The first three models perform word discrimination using DTW on frame-level embeddings of word segments; model 1 works directly on acoustic features, while models 2 and 3 work on features optimized for word discrimination.
Model 3 yields the best previously reported result on this task.
Model 4 is the best acoustic word embedding approach from~\cite{levin+etal_asru13} (Section~\ref{sec:reference_vector}), representing the best previous result for an approach that produces embeddings of whole word~segments.

Models 5 to 11 are the neural
network-based approaches. The effect of using the convolutional layers is evident from the 
large improvement in AP of model 6 over model 5.
Both of these models are trained on the word type labels $\mathcal{W}_{\text{train}}$, which is also the type of supervision used for model 4, making the improvement of model 6 over model 4 noteworthy.\footnote{For model 6, embeddings are taken from the final softmax output. We also experimented with embeddings from the softmax layer but before applying the exponential normalization; this gave worse development results.}
The dimensionality of the acoustic embeddings of model 6, however, is much larger than that of model 4.
We therefore also trained a version of model 6 where the embedding is obtained from a linear bottleneck layer inserted just before the final softmax layer.
The lower-dimensional embeddings from this approach (model 7) still improves on model 4 by 
a sizable margin.

Of the Siamese CNNs, the model with the 
$l_\text{cos hinge}$ loss (model~9, Section~\ref{sec:word_embeddings}) 
outperforms its $l_{\text{cos cos}^2}$ counterpart, and yields a 
large improvement over model 4, which was the previous best acoustic embedding approach.
It also gives similar performance to the 
word classification CNN (model 6), 
even though the pair-wise side information $\mathcal{S}_{\textrm{train}}$ used for model 9 is a weaker form of supervision  than the fully labelled supervision $\mathcal{W}_{\text{train}}$ used in model 6.
When reducing the embedding dimensionality to 50 (model 10), AP is still higher than any of the $d = 50$ competitors.
Model 9's 
improvement over model 3 is also interesting since the former does not use any DTW alignment information. 
Finally, model 11 shows that LDA on the output of model 9 does not yield any improvement, but does produce a much smaller embedding without loss in performance.
This model uses exactly the same word class supervision $\mathcal{W}_{\text{train}}$ as models 6~and~7.

\vspace*{-0.4\baselineskip}
\subsection{Further discussion and analysis}

\begin{figure}[!t]
    \centering
    \includegraphics[width=0.95\linewidth]{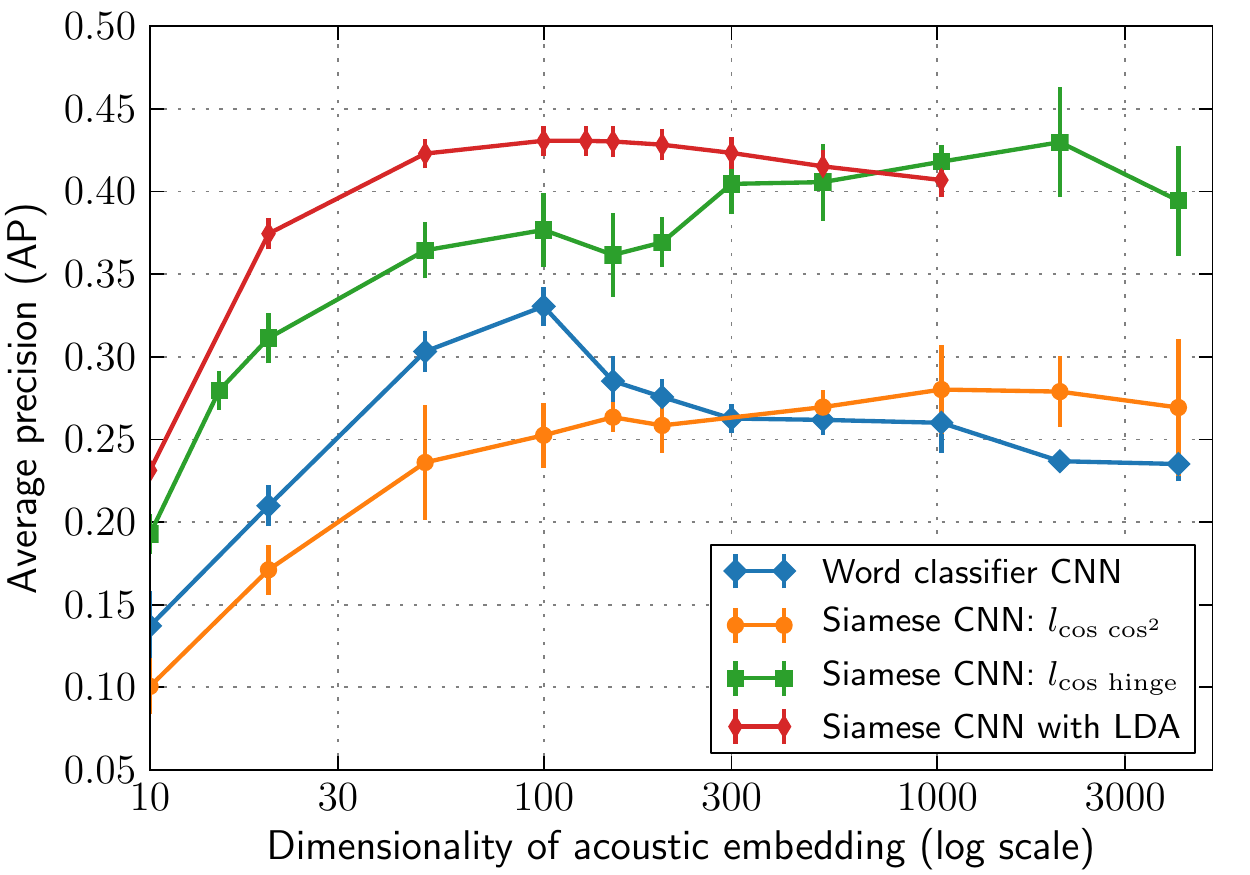}
    \vspace*{-1.0\baselineskip}
    \caption{Average precision (AP) on the development set for different CNN embedding approaches when varying the target dimensionality.}
    \label{fig:cnns_dev}
    \vspace*{-0.5\baselineskip}
\end{figure}

Although the structures of models 8 and 9 are identical, the model using $l_\text{cos hinge}$ 
significantly outperforms its counterpart using $l_{\text{cos cos}^2}$.
This is in line with the fact that a loss like $l_\text{cos hinge}$, which
optimizes embeddings based on \textit{relative} distances between positive and negative pairs, is much more closely 
aligned with the discrimination task than a loss like $l_{\text{cos cos}^2}$, which
looks at distances of word pairs in isolation (without regard to their distances relative to other pairs).
The $l_\text{cos hinge}$ loss also allows more freedom in the model since it does not penalize 
same-word pairs $(\vec{x}_1, \vec{x}_2)$ if they are already more similar by the margin $m$ than the 
corresponding
different-word pairs~$(\vec{x}_1, \vec{x}_3)$.

The closer match between the same-different task and the training loss $l_\text{cos hinge}$ could also explain the improvements over the DTW-based model 3 (both using exactly the same supervision $\mathcal{S}_\text{train}$); this latter model aims to learn better features at the local 
frame level, but does so without regard to the (relative) similarities of complete segments.

Figure~\ref{fig:cnns_dev} shows AP on the development set when varying the target dimensionalities of the different CNN-based approaches (see Section~\ref{sec:dimred}).
The $l_\text{cos hinge}$ Siamese CNN outperforms all the other models (apart from the post-processed LDA model) at all operating points, and gives stable performance over a range of dimensionalities ($300$ and onwards).
The word classifier CNN does much worse in this case (compared to the result of model 6),
perhaps since embeddings here are not taken from the final layer, which is explicitly optimized for word classification, but 
from an intermediate layer.
In contrast, for the Siamese CNNs, embeddings are always obtained directly from the layer that is optimized in the target loss.
The figure also shows that when word labels $\mathcal{W}_\text{train}$ are available, compact embeddings can be obtained by performing LDA on top of the Siamese CNN representation, without loss in performance; this can prove to be important for downstream tasks which might require smaller embeddings.

Here, a relatively small set of labelled word examples $\mathcal{Y}_\text{train}$ is used to 
train the word classifier networks (as also done in the studies we compare to).
In contrast, by using pairs of words and relative comparisons between them,
a much larger~set~$\mathcal{S}_\text{train}$ is used for training Siamese networks.
This type of paired supervision is~ideal~for generalizing to unseen word types, and is often easier to obtain in low-resource settings (see Section~\ref{sec:word_embeddings}).
While frame-level feature learning (model 3) can also use the larger pair-wise training set $\mathcal{S}_\text{train}$, 
such approaches need to be coupled with DTW, which is~limiting.

%% file: conc.tex
\section{Conclusion}

We studied several acoustic word embedding approaches based on convolutional neural networks (CNNs); these networks take a whole-word speech segment as input and produce a fixed-dimensional vector.
Our best new approach is a Siamese CNN that uses a hinge-based loss function to minimize the distance between word pairs of the same type relative to the distance between pairs of different types.
On the \textit{same-different} word discrimination task, this approach yields an average precision (AP) of 0.549,  an improvement over the best previously published results on this task with whole-word embeddings (0.365 AP) and DTW with learned frame features (0.469 AP).
A word classifier CNN performs similarly (0.532 AP) to the Siamese CNN, but requires much stronger labelled supervision, and performs worse at smaller dimensionalties.
Future work will consider sequence models (e.g.\ RNNs, LSTMs), and will apply these embeddings to downstream tasks such as term discovery,
speech recognition, and~search.

%% file: main.bbl
\begin{thebibliography}{10}

\bibitem{dewachter+etal_taslp07}
M.~De~Wachter, M.~Matton, K.~Demuynck, P.~Wambacq, R.~Cools, and
  D.~Van~Compernolle,
\newblock ``Template-based continuous speech recognition,''
\newblock {\em IEEE Trans. Audio, Speech, Language Process.}, vol. 15, no. 4,
  pp. 1377--1390, 2007.

\bibitem{heigold+etal_icassp12}
G.~Heigold, P.~Nguyen, M.~Weintraub, and V.~Vanhoucke,
\newblock ``Investigations on exemplar-based features for speech recognition
  towards thousands of hours of unsupervised, noisy data,''
\newblock in {\em Proc. ICASSP}, 2012.

\bibitem{levin+etal_asru13}
K.~Levin, K.~Henry, A.~Jansen, and K.~Livescu,
\newblock ``Fixed-dimensional acoustic embeddings of variable-length segments
  in low-resource settings,''
\newblock in {\em Proc. ASRU}, 2013.

\bibitem{bengio+heigold_interspeech14}
S.~Bengio and G.~Heigold,
\newblock ``Word embeddings for speech recognition,''
\newblock in {\em Proc. Interspeech}, 2014.

\bibitem{guoguo+etal_icassp15}
G.~Chen, C.~Parada, and T.~N. Sainath,
\newblock ``Query-by-example keyword spotting using long short-term memory
  networks,''
\newblock in {\em Proc. ICASSP}, 2015.

\bibitem{maas+etal_icmlwrl12}
A.~L. Maas, S.~D. Miller, T.~M. O'neil, A.~Y. Ng, and P.~Nguyen,
\newblock ``Word-level acoustic modeling with convolutional vector
  regression,''
\newblock in {\em Proc. ICML Workshop Representation Learn.}, 2012.

\bibitem{rasanen_cogsci15}
O.~J. {R{\"a}s{\"a}nen},
\newblock ``Generating hyperdimensional distributed representations from
  continuous-valued multivariate sensory input,''
\newblock in {\em Proc. {CogSci}}, 2015.

\bibitem{myers+rabiner_tassp81}
C.~S. Myers and L.~R. Rabiner,
\newblock ``Connected digit recognition using a level-building {DTW}
  algorithm,''
\newblock {\em IEEE Trans. Acoust., Speech, Signal Process.}, vol. 29, no. 3,
  pp. 351--363, 1981.

\bibitem{zhang+glass_asru09}
Y.~Zhang and J.~R. Glass,
\newblock ``Unsupervised spoken keyword spotting via segmental {DTW} on
  {G}aussian posteriorgrams,''
\newblock in {\em Proc. ASRU}, 2009.

\bibitem{zhang+etal_icassp12}
Y.~Zhang, R.~Salakhutdinov, H.-A. Chang, and J.~R. Glass,
\newblock ``Resource configurable spoken query detection using deep {B}oltzmann
  machines,''
\newblock in {\em Proc. ICASSP}, 2012.

\bibitem{jansen+etal_icassp13b}
A.~Jansen, S.~Thomas, and H.~Hermansky,
\newblock ``Weak top-down constraints for unsupervised acoustic model
  training,''
\newblock in {\em Proc. ICASSP}, 2013.

\bibitem{kamper+etal_icassp15}
H.~Kamper, M.~Elsner, A.~Jansen, and S.~J. Goldwater,
\newblock ``Unsupervised neural network based feature extraction using weak
  top-down constraints,''
\newblock in {\em Proc. ICASSP}, 2015.

\bibitem{thiolliere+etal_interspeech15}
R.~Thiolli{\`e}re, E.~Dunbar, G.~Synnaeve, M.~Versteegh, and E.~Dupoux,
\newblock ``A hybrid dynamic time warping-deep neural network architecture for
  unsupervised acoustic modeling,''
\newblock in {\em Proc. Interspeech}, 2015.

\bibitem{rabiner+etal_tassp78}
L.~R. Rabiner, A.~E. Rosenberg, and S.~E. Levinson,
\newblock ``Considerations in dynamic time warping algorithms for discrete word
  recognition,''
\newblock {\em IEEE Trans. Acoust., Speech, Signal Process.}, vol. 26, no. 6,
  pp. 575--582, 1978.

\bibitem{levin+etal_icassp15}
K.~Levin, A.~Jansen, and B.~Van~Durme,
\newblock ``Segmental acoustic indexing for zero resource keyword search,''
\newblock in {\em Proc. ICASSP}, 2015.

\bibitem{carlin+etal_icassp11}
M.~A. Carlin, S.~Thomas, A.~Jansen, and H.~Hermansky,
\newblock ``Rapid evaluation of speech representations for spoken term
  discovery,''
\newblock in {\em Proc. Interspeech}, 2011.

\bibitem{bromley+etal_ijpr93}
J.~Bromley, J.~W. Bentz, L.~Bottou, I.~Guyon, Y.~LeCun, C.~Moore,
  E.~S{\"a}ckinger, and R.~Shah,
\newblock ``Signature verification using a `{Siamese}' time delay neural
  network,''
\newblock {\em Int. J. Pattern Rec.}, vol. 7, no. 4, pp. 669--688, 1993.

\bibitem{kamper2015fully}
H.~Kamper, A.~Jansen, and S.~Goldwater,
\newblock ``Fully unsupervised small-vocabulary speech recognition using a
  segmental {Bayesian} model,''
\newblock in {\em Proc. Interspeech}, 2015.

\bibitem{renshaw+etal_interspeech15}
D.~Renshaw, H.~Kamper, A.~Jansen, and S.~J. Goldwater,
\newblock ``A comparison of neural network methods for unsupervised
  representation learning on the {Zero Resource Speech Challenge},''
\newblock in {\em Proc. Interspeech}, 2015.

\bibitem{park+glass_taslp08}
A.~S. Park and J.~R. Glass,
\newblock ``Unsupervised pattern discovery in speech,''
\newblock {\em IEEE Trans. Audio, Speech, Language Process.}, vol. 16, no. 1,
  pp. 186--197, 2008.

\bibitem{jansen+vandurme_asru11}
A.~Jansen and B.~Van~Durme,
\newblock ``Efficient spoken term discovery using randomized algorithms,''
\newblock in {\em Proc. ASRU}, 2011.

\bibitem{badino+etal_icassp14}
L.~Badino, C.~Canevari, L.~Fadiga, and G.~Metta,
\newblock ``An auto-encoder based approach to unsupervised learning of subword
  units,''
\newblock in {\em Proc. ICASSP}, 2014.

\bibitem{huang+etal_cimk13}
P.-S. Huang, X.~He, J.~Gao, L.~Deng, A.~Acero, and L.~Heck,
\newblock ``Learning deep structured semantic models for web search using
  clickthrough data,''
\newblock in {\em Proc. CIMK}, 2013.

\bibitem{mikolov+etal_arxiv13}
T.~Mikolov, K.~Chen, G.~Corrado, and J.~Dean,
\newblock ``Efficient estimation of word representations in vector space,''
\newblock {\em arXiv preprint arXiv:1301.3781}, 2013.

\bibitem{wieting+etal_tacl15}
J.~Wieting, M.~Bansal, K.~Gimpel, and K.~Livescu,
\newblock ``From paraphrase database to compositional paraphrase model and
  back,''
\newblock {\em Trans. ACL}, vol. 3, pp. 345--358, 2015.

\bibitem{hadsell+etal_cvpr06}
R.~Hadsell, S.~Chopra, and Y.~LeCun,
\newblock ``Dimensionality reduction by learning an invariant mapping,''
\newblock in {\em Proc. CVPR}, 2006.

\bibitem{synnaeve+etal_slt14}
G.~Synnaeve, T.~Schatz, and E.~Dupoux,
\newblock ``Phonetics embedding learning with side information,''
\newblock in {\em Proc. SLT}, 2014.

\bibitem{bergstra+etal_scipy10}
J.~Bergstra, O.~Breuleux, F.~Bastien, P.~Lamblin, R.~Pascanu, G.~Desjardins,
  J.~Turian, D.~Warde-Farley, and Y.~Bengio,
\newblock ``Theano: a {CPU} and {GPU} math expression compiler,''
\newblock in {\em Proc. SciPy}, 2010.

\bibitem{zeiler_arxiv12}
M.~D. Zeiler,
\newblock ``{ADADELTA}: An adaptive learning rate method,''
\newblock {\em arXiv preprint arXiv:1212.5701}, 2012.

\end{thebibliography}
